\documentclass[10pt,twocolumn,letterpaper]{article}
\usepackage{multirow}

\usepackage[pagenumbers]{cvpr} 
\usepackage[dvipsnames]{xcolor}

\definecolor{cvprblue}{rgb}{0.21,0.49,0.74}
\usepackage[pagebackref,breaklinks,colorlinks,citecolor=cvprblue]{hyperref}
\usepackage[accsupp]{axessibility} 
\usepackage{graphicx} 
\usepackage{subcaption} 

\def\paperID{5656} 
\def\confName{CVPR}
\def\confYear{2024}

\newcommand{\mytabular}[2]{\centering\scalebox{#1}{#2}}

\title{Towards Automated Movie Trailer Generation}

\author{
Dawit Mureja Argaw$^{1,2}$ \quad
Mattia Soldan$^1$ \quad
Alejandro Pardo$^1$ \quad
Chen Zhao$^1$\thanks{Corresponding author.} \quad \\
Fabian Caba Heilbron$^3$ \quad
Joon Son Chung$^2$ \quad 
Bernard Ghanem$^1$
 \\ \\
$^1$ King Abdullah University of Science and Technology (KAUST) \quad $^2$  KAIST \quad $^3$ Adobe Research\\
}

\begin{document}
%
%
%
%
%
%
%
\newcommand{\ba}{{\mathbf{a}}}
\newcommand{\bb}{{\mathbf{b}}}
\newcommand{\bc}{{\mathbf{c}}}
\newcommand{\bd}{{\mathbf{d}}}
\newcommand{\bolde}{{\mathbf{e}}}
\newcommand{\boldf}{{\mathbf{f}}}
\newcommand{\bg}{{\mathbf{g}}}
\newcommand{\bh}{{\mathbf{h}}}
\newcommand{\bi}{{\mathbf{i}}}
\newcommand{\bj}{{\mathbf{j}}}
\newcommand{\bk}{{\mathbf{k}}}
\newcommand{\bl}{{\mathbf{l}}}
\newcommand{\bm}{{\mathbf{m}}}
\newcommand{\bn}{{\mathbf{n}}}
\newcommand{\bo}{{\mathbf{o}}}
\newcommand{\bp}{{\mathbf{p}}}
\newcommand{\bq}{{\mathbf{q}}}
\newcommand{\br}{{\mathbf{r}}}
\newcommand{\bs}{{\mathbf{s}}}
\newcommand{\bt}{{\mathbf{t}}}
\newcommand{\bu}{{\mathbf{u}}}
\newcommand{\bv}{{\mathbf{v}}}
\newcommand{\bw}{{\mathbf{w}}}
\newcommand{\bx}{{\mathbf{x}}}
\newcommand{\by}{{\mathbf{y}}}
\newcommand{\bz}{{\mathbf{z}}}
\newcommand{\model}{{MODEL}}

\newcommand{\bA}{\mathbf{A}}
\newcommand{\bB}{\mathbf{B}}
\newcommand{\bC}{\mathbf{C}}
\newcommand{\bD}{\mathbf{D}}
\newcommand{\bE}{\mathbf{E}}
\newcommand{\bF}{\mathbf{F}}
\newcommand{\bG}{\mathbf{G}}
\newcommand{\bH}{\mathbf{H}}
\newcommand{\bI}{\mathbf{I}}
\newcommand{\bJ}{\mathbf{J}}
\newcommand{\bK}{\mathbf{K}}
\newcommand{\bL}{\mathbf{L}}
\newcommand{\bM}{\mathbf{M}}
\newcommand{\bN}{\mathbf{N}}
\newcommand{\bO}{\mathbf{O}}
\newcommand{\bP}{\mathbf{P}}
\newcommand{\bQ}{\mathbf{Q}}
\newcommand{\bR}{\mathbf{R}}
\newcommand{\bS}{\mathbf{S}}
\newcommand{\bT}{\mathbf{T}}
\newcommand{\bU}{\mathbf{U}}
\newcommand{\bV}{\mathbf{V}}
\newcommand{\bW}{\mathbf{W}}
\newcommand{\bX}{\mathbf{X}}
\newcommand{\bY}{\mathbf{Y}}
\newcommand{\bZ}{\mathbf{Z}}

\newcommand{\calA}{{\mathcal{A}}}
\newcommand{\calB}{{\mathcal{B}}}
\newcommand{\calC}{{\mathcal{C}}}
\newcommand{\calD}{{\mathcal{D}}}
\newcommand{\calE}{{\mathcal{E}}}
\newcommand{\calF}{{\mathcal{F}}}
\newcommand{\calG}{{\mathcal{G}}}
\newcommand{\calH}{{\mathcal{H}}}
\newcommand{\calI}{{\mathcal{I}}}
\newcommand{\calJ}{{\mathcal{J}}}
\newcommand{\calK}{{\mathcal{K}}}
\newcommand{\calL}{{\mathcal{L}}}
\newcommand{\calM}{{\mathcal{M}}}
\newcommand{\calN}{{\mathcal{N}}}
\newcommand{\calO}{{\mathcal{O}}}
\newcommand{\calP}{{\mathcal{P}}}
\newcommand{\calQ}{{\mathcal{Q}}}
\newcommand{\calR}{{\mathcal{R}}}
\newcommand{\calS}{{\mathcal{S}}}
\newcommand{\calT}{{\mathcal{T}}}
\newcommand{\calU}{{\mathcal{U}}}
\newcommand{\calV}{{\mathcal{V}}}
\newcommand{\calW}{{\mathcal{W}}}
\newcommand{\calX}{{\mathcal{X}}}
\newcommand{\calY}{{\mathcal{Y}}}
\newcommand{\calZ}{{\mathcal{Z}}}
\newcommand{\calbX}{\mbox{\boldmath $\mathcal{X}$}}
\newcommand{\calbY}{\mbox{\boldmath $\mathcal{Y}$}}

\newcommand{\bcalA}{\mbox{\boldmath $\calA$}}
\newcommand{\bcalB}{\mbox{\boldmath $\calB$}}
\newcommand{\bcalC}{\mbox{\boldmath $\calC$}}
\newcommand{\bcalD}{\mbox{\boldmath $\calD$}}
\newcommand{\bcalE}{\mbox{\boldmath $\calE$}}
\newcommand{\bcalF}{\mbox{\boldmath $\calF$}}
\newcommand{\bcalG}{\mbox{\boldmath $\calG$}}
\newcommand{\bcalH}{\mbox{\boldmath $\calH$}}
\newcommand{\bcalI}{\mbox{\boldmath $\calI$}}
\newcommand{\bcalJ}{\mbox{\boldmath $\calJ$}}
\newcommand{\bcalK}{\mbox{\boldmath $\calK$}}
\newcommand{\bcalL}{\mbox{\boldmath $\calL$}}
\newcommand{\bcalM}{\mbox{\boldmath $\calM$}}
\newcommand{\bcalN}{\mbox{\boldmath $\calN$}}
\newcommand{\bcalO}{\mbox{\boldmath $\calO$}}
\newcommand{\bcalP}{\mbox{\boldmath $\calP$}}
\newcommand{\bcalQ}{\mbox{\boldmath $\calQ$}}
\newcommand{\bcalR}{\mbox{\boldmath $\calR$}}
\newcommand{\bcalS}{\mbox{\boldmath $\calS$}}
\newcommand{\bcalT}{\mbox{\boldmath $\calT$}}
\newcommand{\bcalU}{\mbox{\boldmath $\calU$}}
\newcommand{\bcalV}{\mbox{\boldmath $\calV$}}
\newcommand{\bcalW}{\mbox{\boldmath $\calW$}}
\newcommand{\bcalX}{\mbox{\boldmath $\calX$}}
\newcommand{\bcalY}{\mbox{\boldmath $\calY$}}
\newcommand{\bcalZ}{\mbox{\boldmath $\calZ$}}

\newcommand{\sfA}{\mbox{$\mathsf A$}}
\newcommand{\sfB}{\mbox{$\mathsf B$}}
\newcommand{\sfC}{\mbox{$\mathsf C$}}
\newcommand{\sfD}{\mbox{$\mathsf D$}}
\newcommand{\sfE}{\mbox{$\mathsf E$}}
\newcommand{\sfF}{\mbox{$\mathsf F$}}
\newcommand{\sfG}{\mbox{$\mathsf G$}}
\newcommand{\sfH}{\mbox{$\mathsf H$}}
\newcommand{\sfI}{\mbox{$\mathsf I$}}
\newcommand{\sfJ}{\mbox{$\mathsf J$}}
\newcommand{\sfK}{\mbox{$\mathsf K$}}
\newcommand{\sfL}{\mbox{$\mathsf L$}}
\newcommand{\sfM}{\mbox{$\mathsf M$}}
\newcommand{\sfN}{\mbox{$\mathsf N$}}
\newcommand{\sfO}{\mbox{$\mathsf O$}}
\newcommand{\sfP}{\mbox{$\mathsf P$}}
\newcommand{\sfQ}{\mbox{$\mathsf Q$}}
\newcommand{\sfR}{\mbox{$\mathsf R$}}
\newcommand{\sfS}{\mbox{$\mathsf S$}}
\newcommand{\sfT}{\mbox{$\mathsf T$}}
\newcommand{\sfU}{\mbox{$\mathsf U$}}
\newcommand{\sfV}{\mbox{$\mathsf V$}}
\newcommand{\sfW}{\mbox{$\mathsf W$}}
\newcommand{\sfX}{\mbox{$\mathsf X$}}
\newcommand{\sfY}{\mbox{$\mathsf Y$}}
\newcommand{\sfZ}{\mbox{$\mathsf Z$}}

\newcommand{\balpha}{\mbox{\boldmath $\alpha$}}
\newcommand{\bbeta}{\mbox{\boldmath $\beta$}}
\newcommand{\bgamma}{\mbox{\boldmath $\gamma$}}
\newcommand{\bdelta}{\mbox{\boldmath $\delta$}}
\newcommand{\bepsilon}{\mbox{\boldmath $\epsilon$}}
\newcommand{\bvarepsilon}{\mbox{\boldmath $\varepsilon$}}
\newcommand{\bzeta}{\mbox{\boldmath $\zeta$}}
\newcommand{\boldeta}{\mbox{\boldmath $\eta$}}
\newcommand{\btheta}{\mbox{\boldmath $\theta$}}
\newcommand{\bvartheta}{\mbox{\boldmath $\vartheta$}}
\newcommand{\biota}{\mbox{\boldmath $\iota$}}
\newcommand{\bkappa}{\mbox{\boldmath $\kappa$}}
\newcommand{\blambda}{\mbox{\boldmath $\lambda$}}
\newcommand{\bmu}{\mbox{\boldmath $\mu$}}
\newcommand{\bnu}{\mbox{\boldmath $\nu$}}
\newcommand{\bxi}{\mbox{\boldmath $\xi$}}
\newcommand{\bpi}{\mbox{\boldmath $\pi$}}
\newcommand{\bvarpi}{\mbox{\boldmath $\varpi$}}
\newcommand{\brho}{\mbox{\boldmath $\rho$}}
\newcommand{\bvarrho}{\mbox{\boldmath $\varrho$}}
\newcommand{\bsigma}{\mbox{\boldmath $\sigma$}}
\newcommand{\bvarsigma}{\mbox{\boldmath $\varsigma$}}
\newcommand{\btau}{\mbox{\boldmath $\tau$}}
\newcommand{\bupsilon}{\mbox{\boldmath $\upsilon$}}
\newcommand{\bphi}{\mbox{\boldmath $\phi$}}
\newcommand{\bvarphi}{\mbox{\boldmath $\varphi$}}
\newcommand{\bchi}{\mbox{\boldmath $\chi$}}
\newcommand{\bpsi}{\mbox{\boldmath $\psi$}}
\newcommand{\bomega}{\mbox{\boldmath $\omega$}}

\newcommand{\bGamma}{\mbox{\boldmath $\Gamma$}}
\newcommand{\bDelta}{\mbox{\boldmath $\Delta$}}
\newcommand{\bTheta}{\mbox{\boldmath $\Theta$}}
\newcommand{\bLambda}{\mbox{\boldmath $\Lambda$}}
\newcommand{\bXi}{\mbox{\boldmath $\Xi$}}
\newcommand{\bPi}{\mbox{\boldmath $\Pi$}}
\newcommand{\bSigma}{\mbox{\boldmath $\Sigma$}}
\newcommand{\bUpsilon}{\mbox{\boldmath $\Upsilon$}}
\newcommand{\bPhi}{\mbox{\boldmath $\Phi$}}
\newcommand{\bPsi}{\mbox{\boldmath $\Psi$}}
\newcommand{\bOmega}{\mbox{\boldmath $\Omega$}}

\newcommand{\veca}{{\vec{\ba}}}
\newcommand{\vecb}{{\vec{\bb}}}
\newcommand{\vecc}{{\vec{\bc}}}
\newcommand{\vecd}{{\vec{\bd}}}
\newcommand{\vece}{{\vec{\bolde}}}
\newcommand{\vecf}{{\vec{\boldf}}}
\newcommand{\vecg}{{\vec{\bg}}}
\newcommand{\vech}{{\vec{\bh}}}
\newcommand{\veci}{{\vec{\bi}}}
\newcommand{\vecj}{{\vec{\bj}}}
\newcommand{\veck}{{\vec{\bk}}}
\newcommand{\vecl}{{\vec{\bl}}}
\newcommand{\vecm}{{\vec{\bm}}}
\newcommand{\vecn}{{\vec{\bn}}}
\newcommand{\veco}{{\vec{\bo}}}
\newcommand{\vecp}{{\vec{\bp}}}
\newcommand{\vecq}{{\vec{\bq}}}
\newcommand{\vecr}{{\vec{\br}}}
\newcommand{\vecs}{{\vec{\bs}}}
\newcommand{\vect}{{\vec{\bt}}}
\newcommand{\vecu}{{\vec{\bu}}}
\newcommand{\vecv}{{\vec{\bv}}}
\newcommand{\vecw}{{\vec{\bw}}}
\newcommand{\vecx}{{\vec{\bx}}}
\newcommand{\vecy}{{\vec{\by}}}
\newcommand{\vecz}{{\vec{\bz}}}

\newcommand{\vecxi}{{\vec{\bxi}}}
\newcommand{\vecphi}{{\vec{\bphi}}}
\newcommand{\vecvarphi}{{\vec{\bvarphi}}}
\newcommand{\vecbeta}{{\vec{\bbeta}}}
\newcommand{\vecdelta}{{\vec{\bdelta}}}
\newcommand{\vectheta}{{\vec{\btheta}}}

\newcommand{\Real}{\mathbb R}
\newcommand{\Complex}{\mathbb C}
\newcommand{\Natural}{\mathbb N}
\newcommand{\Integer}{\mathbb Z}


\newcommand{\bone}{\mbox{\boldmath $1$}}
\newcommand{\bzero}{\mbox{\boldmath $0$}}
\newcommand{\0}{{\bf 0}}

\newcommand{\be}{\begin{eqnarray}}
\newcommand{\ee}{\end{eqnarray}}
\newcommand{\bee}{\begin{eqnarray*}}
\newcommand{\eee}{\end{eqnarray*}}

\newcommand{\matrixb}{\left[ \begin{array}}
\newcommand{\matrixe}{\end{array} \right]}

\newcommand{\argmax}{\operatornamewithlimits{\arg \max}}
\newcommand{\argmin}{\operatornamewithlimits{\arg \min}}

\newcommand{\mean}[1]{\left \langle #1 \right \rangle}
\newcommand{\ave}{\mathbb E}
\newcommand{\E}{\mathbb E}
\newcommand{\empha}[1]{{\color{red} \bf #1}}
\newcommand{\fracpartial}[2]{\frac{\partial #1}{\partial  #2}}
\newcommand{\incomplete}[1]{\textcolor{red}{#1}}

\def\doublespace{\renewcommand{\baselinestretch}{2}\large\normalsize}
\def\singlespace{\renewcommand{\baselinestretch}{1}\large\normalsize}
\def\onehalfspace{\renewcommand{\baselinestretch}{1.5}\large\normalsize}
\def\onequaterspace{\renewcommand{\baselinestretch}{1.3}\large\normalsize}
\def\threequaterspace{\renewcommand{\baselinestretch}{1.7}\large\normalsize}
\def\smallspace{\renewcommand{\baselinestretch}{-.9}\large\normalsize}
\def\tinyspace{\renewcommand{\baselinestretch}{-.7}\large\normalsize}

\newcommand{\tr} { \textrm{tr} }
\newcommand{\re} { \textrm{re} }
\newcommand{\im} { \textrm{im} }
\newcommand{\diag} { \textrm{diag} }
\newcommand{\ddiag} { \textrm{ddiag} }
\newcommand{\off} { \textrm{off} }
\newcommand{\vectxt} { \textrm{vec} }

\newcommand{\lla}{\left\langle}
\newcommand{\rra}{\right\rangle}
\newcommand{\llbr}{\left\lbrack}
\newcommand{\rrbr}{\right\rbrack}
\newcommand{\llb}{\left\lbrace}
\newcommand{\rrb}{\right\rbrace}


\newcommand{\RR}{I\!\!R} 
\newcommand{\Nat}{I\!\!N} 
\newcommand{\CC}{I\!\!\!\!C} 

\newcommand{\Tref}[1]{Table~\ref{#1}}
\newcommand{\Eref}[1]{Eq.~(\ref{#1})}
\newcommand{\Fref}[1]{Fig.~\ref{#1}}
\newcommand{\FCref}[1]{Chapter.~\ref{#1}}
\newcommand{\Sref}[1]{Sec.~\ref{#1}}
\newcommand{\Aref}[1]{Algo.~\ref{#1}}

\def\eg{\emph{e.g.}}
\def\Eg{\emph{E.g.}}
\def\etal{\emph{et al.}}
\def\ie{\emph{i.e.}}

\maketitle
\begin{abstract}\label{sec:abstract}
Movie trailers are an essential tool for promoting films and attracting audiences. However, the process of creating trailers can be time-consuming and expensive. To streamline this process, we propose an automatic trailer generation framework that generates plausible trailers from a full movie by automating shot selection and composition. Our approach draws inspiration from machine translation techniques and models the movies and trailers as sequences of shots, thus formulating the trailer generation problem as a sequence-to-sequence task. We introduce Trailer Generation Transformer (TGT), a deep-learning framework utilizing an encoder-decoder architecture. 
TGT movie encoder is tasked with contextualizing each movie shot representation via self-attention, while the autoregressive trailer decoder predicts the feature representation of the next trailer shot, accounting for the relevance of shots' temporal order in trailers. 
Our TGT significantly outperforms previous methods on a comprehensive suite of metrics. 
\end{abstract}
    
\vspace{-4mm}
\section{Introduction}\label{sec:introduction}
\begin{figure}[!t]
    \centering
    \includegraphics[width=\linewidth,trim={0cm 0cm 0cm 0cm},clip]{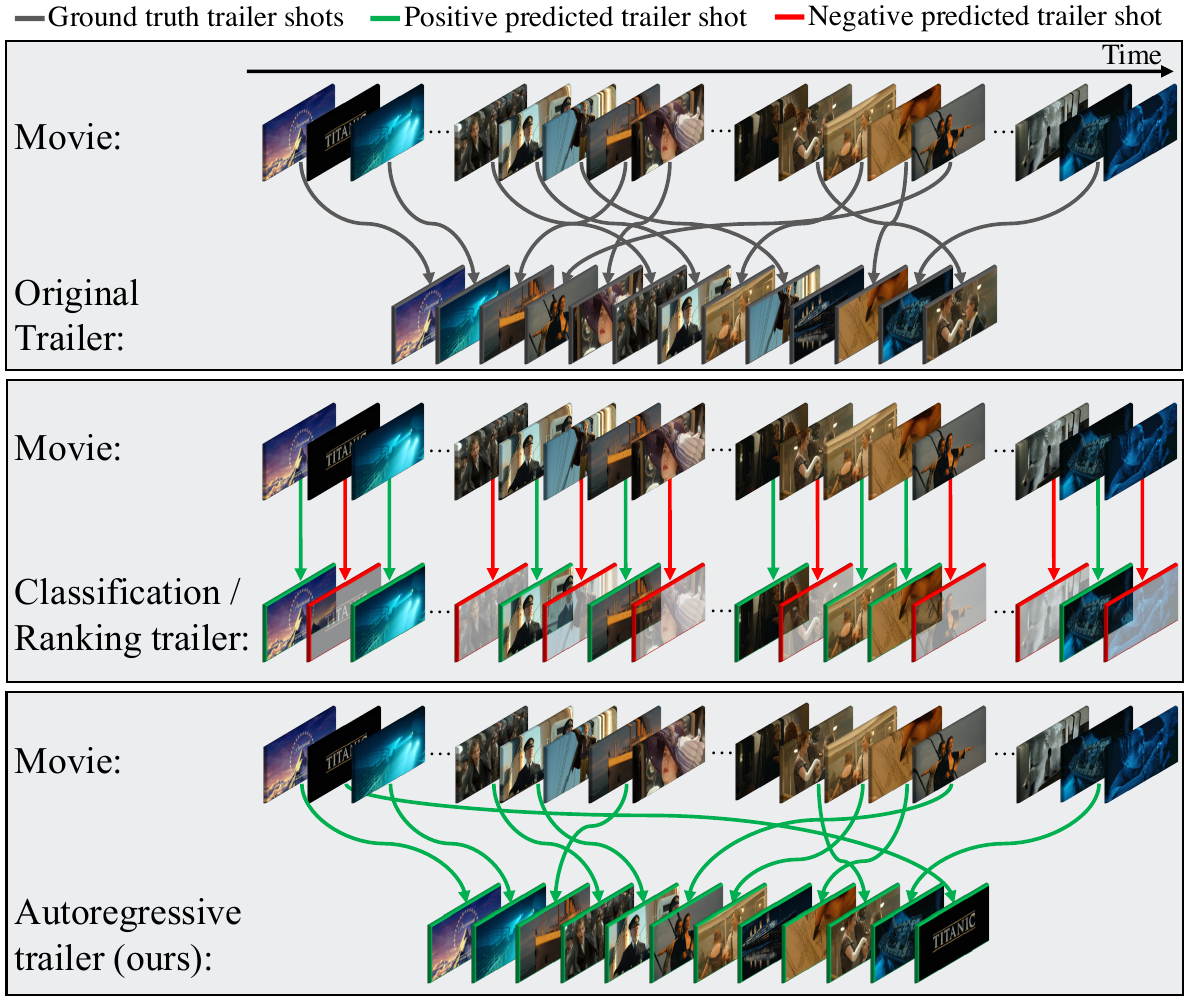}
    \caption{\textbf{Trailer Generation Problem and Solutions.} The top row depicts the movie and expert-created trailer. The process composes shots from the movie in non-chronological order to create a compelling and intriguing story. Depicted in the central row are the classification/ranking strategies that classify/rank each shot in the movie independently (classification) or with limited relative interaction (ranking). The bottom row represents our approach which can reason over the entire input movie sequence before producing a provisional trailer with non-chronological shots order.}
    \label{fig:teaser}
    \vspace{-0.5cm}
\end{figure}

Movie trailers are essential marketing tools for the film industry, generating anticipation by showcasing captivating scenes, storylines, and cast members. They enable studios to fuel marketing campaigns and build interest ahead of a movie's release. For audiences, trailers serve as decision-making tools, offering a condensed preview of the film's content. Despite their importance, creating trailers is costly, time-consuming, and demands expert knowledge. 

The process of creating a trailer can be broadly divided into two stages\footnote{\href{https://www.youtube.com/watch?v=zkEXtwCL684}{https://www.youtube.com/watch?v=zkEXtwCL684}}. In the \textit{first} stage, video editors immerse themselves in the entire movie, viewing all the shots. They carefully select relevant trailer shots and arrange them in a specific order to craft an engaging flow and rhythm for the movie trailer (see~\Fref{fig:teaser}, top row). This is a time-consuming and tedious process as the editor has to sort through an extensive collection of movie shots~\cite{wang2020learning}. The \textit{second} stage involves fine-editing which incorporates dialogue and sound modeling into the trailer. This work delves into the domain of Automatic Trailer Generation (ATG) with the aim of streamlining the first stage, \ie~shot selection and sequencing, to create a trailer montage from a given full movie (see~\Fref{fig:teaser}, bottom row).

ATG has remained a relatively underexplored problem mainly due to the complexity of the task and the lack of well-established benchmarks. Nevertheless, few prior works~\cite{Gaikwad_2021_ICCV, smith2017harnessing, narasimhan2021clip, wang2020learning} have made efforts to tackle this challenging task. These works have approached trailer generation in different ways. Some works~\cite{smith2017harnessing, narasimhan2021clip} have framed it as a binary \textit{classification} problem, where each shot in a movie is classified as either a trailer shot or not, while others~\cite{wang2020learning, Gaikwad_2021_ICCV} have treated it as a \textit{ranking} problem, wherein the top-ranked movie shots are considered as trailer shots. 

While the progress has been promising, previous works on trailer generation exhibit several limitations, mainly due to their problem formulations. First, both classification and ranking are likely vulnerable to a long-tail distribution problem since only a few percent of movie shots are trailer shots, and hence, there will always be a class imbalance in classification~\cite{smith2017harnessing} or a significant amount of hard negatives in ranking~\cite{wang2020learning}. Second, in classification-based works~\cite{smith2017harnessing, narasimhan2021clip}, all trailer shots are classified in parallel, and the decision on whether a given shot is a trailer is not conditioned by shots that are already selected as trailer shots. This eventually leads to numerous repetitive shots being categorized as trailer shots. Similarly, ranking-based works~\cite{wang2020learning, Gaikwad_2021_ICCV} only consider two shots at a time. Third, and most importantly, existing works do not consider shot composition (ordering), as their formulation does not allow it. Consequently, the shot sequence in the generated trailer mirrors the chronological order of the shots in the original movie. See ~\Fref{fig:teaser}, central row, for a graphical illustration of the predicted trailer generated by classification and ranking methods.

Our work addresses these limitations by introducing a new trailer generation framework. First, we formulate trailer generation as a regression problem where we predict continuous feature representations of trailer shots instead of a discrete binary class. This makes our approach less vulnerable to a long-tail distribution problem~\cite{zhang2023deep}. Second, we employ an autoregressive approach in which the prediction of a trailer shot embedding at a specific time step is conditioned not only on the input movie sequence but also on the generated trailer sequence up to the current step. Third, we design a model that takes shot composition into account, and we reinforce this concept through training on a large set of movie-trailer pairs with carefully designed losses. 

In pursuit of these specifications, we introduce Trailer Generation Transformer (TGT), a model designed to automatically generate plausible trailers using full movies as input. Our approach uses an encoder-decoder architecture that models ATG as a sequence-to-sequence (seq2seq) problem~\cite{sutskever2014sequence, vaswani2017attention}, where the input sequence corresponds to the movie sequence, and the target sequence represents the desired trailer. To make this process computationally feasible and follow the natural structure of movies, we utilize a shot boundary detector~\cite{souvcek2020transnet} to \textit{tokenize} movies and trailers and use a state-of-the-art visual encoder (\ie~CLIP~\cite{radford2021learning}) to obtain sequences of visual embeddings.

TGT's encoder comprises two components: \textit{trailerness encoder} and \textit{context encoder}. The \textit{trailerness encoder} ingests the entire movie sequence and estimates the likelihood of each shot being relevant for trailer creation. This step identifies the visually captivating shots often used for trailer generation. The scores are fused with the input sequence, a process we term \textit{trailerness encoding} drawing a parallel to positional encoding in \cite{vaswani2017attention}. 
The \textit{context encoder} consists of a stack of transformer encoder layers, leveraging self-attention mechanisms~\cite{vaswani2017attention} to contextualize each shot representation based on the entire movie sequence, effectively achieving temporal modeling across multiple shots.

The decoder of TGT is an autoregressive model that predicts the feature representation of the next trailer shot. This design choice is instrumental in enabling the model to learn shot composition, specifically, the ordering of shots, differentiating our approach from previous works~\cite{Gaikwad_2021_ICCV, smith2017harnessing, narasimhan2021clip, wang2020learning}. See the comparison between the central and bottom rows in~\Fref{fig:teaser}. At training time, the decoder takes the encoder output and the target trailer sequence and employs a causal mask to learn to predict the next shot representation. At test time, the model generates outputs in an autoregressive manner.
Finally, a greedy algorithm based on nearest neighbor retrieval matches the generated trailer shot representations with all shots in the movie, thereby selecting the most suitable shots for assembling the output trailer. 

We curate an extensive dataset of movie-trailer pairs and employ tailored loss functions to train TGT for automatic trailer generation effectively. Furthermore, to foster further research on ATG, our work presents new benchmarks based on two widely-used movie datasets: MAD~\cite{soldan2022mad} and MovieNet~\cite{huang2020movienet}. We construct these benchmarks by pairing corresponding trailers and segmenting both movies and trailers into shot sequences, resulting in two datasets with movie-trailer pairs. We perform extensive experiments to assess the performance of our approach and compare it to previous methods using metrics that redefine how we measure trailer generation across various aspects.

Our contributions are: \textbf{(1)} A novel trailer generation formulation that effectively overcomes the inherent limitations of previous approaches. \textbf{(2)} Trailer Generation Transformer (TGT), an autoregressive encoder-decoder architecture that generates provisional trailers from full movies, which can be refined further by experts. \textbf{(3)} New ATG benchmarks built atop two movie datasets: MAD~\cite{soldan2022mad} and MovieNet~\cite{huang2020movienet}. We evaluate trailer generation performance using a comprehensive suite of metrics that consider all relevant aspects of the process.

\section{Related Works}\label{sec:related_work}

\paragraph{Trailer Generation}
Several papers have attempted to automate the process of trailer generation. Chen~\etal~\cite{chen2004action} analyze movie composition based on sets of rules and grammar to generate a trailer by concatenating shots with a specific ``movie tempo'' above a threshold. Smeaton~\etal~\cite{smeaton2006automatically} employ audiovisual features and a support vector machine to automatically select shots for trailers, highlighting the importance of shot sequence but leaving it to artists. Some works~\cite{Gaikwad_2021_ICCV, hesham2018smart, kawai2007automated} utilize text features from TV show descriptions, subtitles, or metadata, to find similar moments in the video transcript or subtitles and select corresponding shots for trailer creation. Others~\cite{ king2021user, liu2015semi, smith2017harnessing} use audio-visual features to identify potential trailer moments from long videos. 

Some works~\cite{irie2010automatic, mishra2022semi} tackle the trailer generation task using predefined templates and automatically fill them with clips from the initial sequence. In contrast, our approach solely relies on the movie sequence and does not use templates, additional metadata, or audio features to generate the trailer. Finally, Wang~\etal~\cite{wang2020learning} recently proposed CCANet, a deep learning model that utilizes co-attention and contrastive attention modules to match and distinguish trailer moments from non-trailer moments. However, CCANet is limited to a single genre and requires genre-specific training. Our work, on the other hand, is genre-agnostic and operates effectively across the diverse range of genres present in MovieNet \cite{huang2020movienet} and MAD \cite{soldan2022mad}.
\vspace{-3mm}
\paragraph{Video Summarization}
Video summarization aims to select the most important clips from a given video to create a concise video summary. Similar to trailer generation, this task involves a shot selection aspect that models must address. Earlier works attempted video summarization across different video domains without explicit supervision \cite{badamdorj2022contrastive, khosla2013large, lu2013story, mahasseni2017unsupervised, panda2017collaborative, potapov2014category, zhou2018deep}. Other works approached this task using supervised learning, coming from web video summaries~\cite{gygli2014creating} or TV series summaries~\cite{song2015tvsum}. Representative works employed various techniques, including fully convolutional sequence networks that learn from paired data~\cite{rochan2018video} and unpaired~data~\cite{rochan2019video}, graph modeling~\cite{park2020sumgraph}, and determinantal point process (DPP) for structured predictions of video sequences~\cite{li2018local, zhang2016summary, zhang2016video}. 

Recent attempts include using attention-based encoding to score the importance of each frame using regression~\cite{fajtl2019summarizing} or modeling the interaction between video and text to compute a saliency score \cite{xu2023mh}. Additionally, Zhang~\etal~\cite{zhang2018retrospective} proposed a sequence-to-sequence modeling approach using an encoder-decoder LSTM-based architecture to preserve the video semantics in the output sequence. Gan~\etal~\cite{gan2023collaborative} propose leveraging movie trailers as supervision for video summarization, while our work uses pairs of movies and trailers for trailer generation rather than summarization. Lastly, CLIP-It~\cite{narasimhan2021clip} guides video summarization with text through dense video captions, while TL:DW?~\cite{narasimhan2022tl} focuses on exploring cross-modal saliency between video and text (transcript) signals to summarize instructional videos. 

\vspace{-1mm}
\section{Methodology}\label{sec:methodology}
\paragraph{Problem Formulation} Given a movie sequence $\calM$, the problem of automatic trailer generation (ATG) aims at generating its corresponding trailer sequence ${\calT}$\!.\ Here, $\calM$ denotes a sequence of movie shots $\{U_1, U_2,\ldots, U_n\}$ and ${\calT}$\! denotes a sequence of trailer shots $\{V_1, V_2,\ldots, V_m\}$, where $n$ and $m$ represent the number of shots in $\calM$ and $\calT$, respectively. During training, we use paired movie and trailer sequences, denoted as $\{\calM_i, \calT_i\}$, where $i$ refers to the index of the movie-trailer pair. These pairs serve as a basis for the model to learn the essential features and patterns necessary for generating trailers from movie sequences.
In contrast, during testing, the model is given only the movie sequences as input, without any corresponding trailer sequences. The purpose of this setup is to evaluate the model's ability to generate trailer sequences $\mathcal{T'}$ that are coherent and relevant to the given movie sequences $\mathcal{M}$.  
The performance of the model is assessed by comparing the generated trailer sequences $\mathcal{T'}$ to the ground truth trailer sequences $\mathcal{T}$, using suitable evaluation metrics. 
\begin{figure*}[!t]
    \vspace{-0.5cm}
    \centering
    \includegraphics[width=\linewidth,trim={0cm 0cm 0cm 0cm},clip]{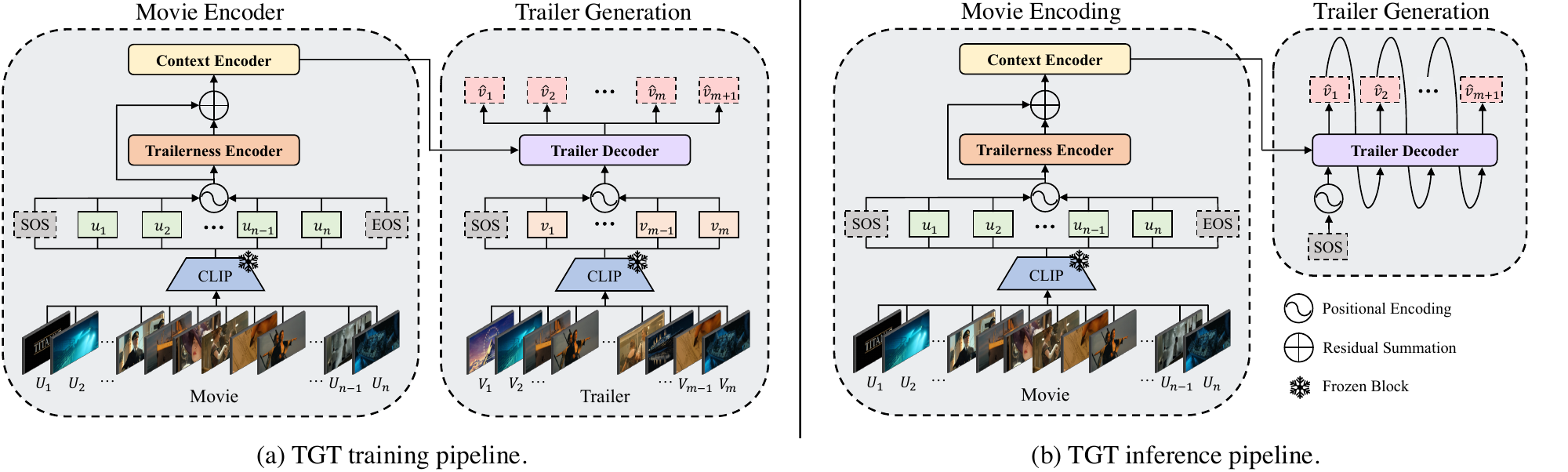}
    \vspace{-0.5cm}
    \caption{
    \textbf{Architecture Overview.} Subfigure (a) illustrates our TGT model's training pipeline. Movies are segmented into shots and transformed into visual embeddings via a pre-trained CLIP model~\cite{radford2021learning}. Enhanced with positional embeddings and trailerness scores, these tokens undergo context encoding. The trailer decoder, during training, uses ground-truth trailer shots as queries for cross-attention with encoder output, then parallelly regresses the next shot feature using a causal mask. Subfigure (b) shows the inference pipeline where the trailer decoder sequentially generates trailer shots in an autoregressive manner while the movie encoder process remains unchanged.
    }
    \label{fig:overview}
    \vspace{-0.4cm}
\end{figure*}
\subsection{Proposed Trailer Generation Transformer}
Camera shots in movies and trailers provide a convenient video structure that we leverage. Our initial step involves segmenting the movie and trailer into shot sequences ($\{U_i\}_{i=1}^{n}$ and $\{V_j\}_{j=1}^{m}$, respectively) using a state-of-the-art shot-boundary detector~\cite{souvcek2020transnet}. 
Learning directly from the pixel space of long-form videos in an end-to-end manner poses demanding computational burdens. Therefore, to efficiently represent the input sequences, we leverage a pretrained CLIP model~\cite{radford2021learning} as a base encoder to extract the feature representations of each shot, thereby transforming the movie and trailer into a sequence of \textit{visual tokens}, \ie~$\{u_i\}_{i=1}^{n}$ and $\{v_j\}_{j=1}^{m}$. To indicate the beginning and end of a movie/trailer sequence, we use learnable start-of-sequence (\texttt{SOS}) and end-of-sequence (\texttt{EOS}) tokens, respectively, as shown in \Fref{fig:overview}.

We draw inspiration from machine translation techniques~\cite{bahdanau2014neural,sutskever2014sequence,vaswani2017attention}. Specifically, we approach the trailer generation problem as a sequence-to-sequence (seq2seq) task~\cite{sutskever2014sequence, vaswani2017attention}. In seq2seq, an encoder captures semantic and syntactic information from an input token sequence and converts it into a contextualized sequence. Then, a decoder generates the translated sequence by producing tokens one at a time, conditioned on the contextualized sequence and previously generated words.
We apply this concept to model the automatic trailer generation (ATG) problem, essentially ``translating'' a movie into a trailer. To do this, we use an attention-based encoder-decoder architecture~\cite{vaswani2017attention} and introduce our Trailer Generation Transformer (TGT) model. TGT consists of two main components: the Movie Encoder (detailed in \Sref{sec:encoder}) and an auto-regressive Trailer Decoder (described in \Sref{sec:decoder}).
\vspace{-3mm}
\subsubsection{Movie Encoder} \label{sec:encoder}
\vspace{-1mm}
The movie encoder aims at encoding a given movie sequence into a contextualized feature sequence by using a trailerness encoder followed by a context encoder, as shown in the Movie Encoder block of~\Fref{fig:overview}.

\vspace{-3mm}
\paragraph{Trailerness Encoder}
Given an input sequence of visual tokens representing a movie, $\{\mathrm{\texttt{SOS}}, u_1,\ldots, u_n,\mathrm{\texttt{EOS}}\}$, a positional encoding layer is first used to embed information about the relative position of the input tokens. We then feed the sequence into a trailerness encoder $\calE_\calT$. The purpose of the trailerness encoder is to observe the full movie and reason over the likeliness of each movie shot to be in a trailer. 
In $\calE_\calT$, a Transformer~\cite{vaswani2017attention} self-attention layer $\calA$, followed by a linear layer $f$ and a Sigmoid layer $s$, is used to predict a trailerness score for each visual token in the input sequence, as formulated below:
\begin{equation}
\small
 \big\{t^\mathrm{p}_i\big\}_{i=0}^{n+1} = s(f(\calA (\texttt{SOS} + \sigma_0, u_1+\sigma_1, \ldots, \texttt{EOS}+\sigma_{n+1}))),
\label{eqn: trailerness}
\end{equation}
where $\sigma_i$ denotes the positional encoding value at position $i$ and $t^\mathrm{p}_i\in [0,1]$ represents the predicted score of a movie shot $U_i$ as a noteworthy shot to be included in a trailer. 
During training, we compute the ground truth trailerness score $t^\mathrm{GT}_i$ for each movie shot by taking the maximum value of the cosine similarity between the movie shot and all the trailer shots (see \Eref{eqn: trailerness_gt}). A high value of $t^\mathrm{GT}_i$ means that a movie shot $U_i$ is in a trailer, whereas a low value indicates that the movie shot is not similar to any shots in the trailer. We optimize the mean-squared error between the predicted and ground truth trailerness scores during training to guide the trailerness encoder $\calE_\calT$ as shown below
in \Eref{eqn: loss_trailerness}
\begin{equation}
 t^\mathrm{GT}_i = \max_j~s_{i,j}, ~~\mathrm{for}~1\le i\le n, \mathrm{where}~s_{i,j} = \frac{u_i\cdot v_j}{\|u_i\| \|v_j\|},
 \label{eqn: trailerness_gt}
\end{equation}
\vspace{-3mm}
\begin{equation}
\calL_t = \sum_{i=0}^{n+1} {|t_i^\mathrm{p} - t_i^\mathrm{GT}|}^2
    \label{eqn: loss_trailerness},
\end{equation}
where $\calL_t$ denotes trailerness loss and the ground truth trailerness score for \texttt{SOS} $(t^\mathrm{GT}_0)$ and \texttt{EOS} $(t^\mathrm{GT}_{n+1})$ tokens is set to 0. After positional encoding, the predicted trailerness scores are added to inject information about the relative degree of trailerness of the movie shots in the sequence (see~\Fref{fig:overview} Movie Encoder).
\vspace{-3mm}
\paragraph{Context Encoder} Trailerness encoding adjusts each positionally encoded visual token by a unique float number, representing the trailerness score. Yet, to understand the relationships between shots in the movie sequence, the trailerness encoded sequence is passed through a context encoder $\calE_\calC$ (see \Eref{eqn:context}). 
The purpose of the context encoder is to obtain a contextualized representation of the input sequence, essential for decoding a trailer. We use a stack of Transformer~\cite{vaswani2017attention} encoder layers for $\calE_\calC$. Through the multi-head self-attention mechanism, $\calE_\calC$ considers the context in which each movie shot appears, generating a representation that encapsulates the meaning of the entire movie sequence.
\vspace{-3mm}
\begin{equation}
    C = \calE_\calC (\texttt{SOS} + \sigma_0 + t_0^\mathrm{p}, 
    u_1+\sigma_1+t_1^\mathrm{p}, 
    \ldots,\texttt{EOS}+\sigma_{n+1}+t_{n+1}^\mathrm{p}),
    \label{eqn:context}
\end{equation}
where $C$ denotes a contextualized movie sequence representation from $\calE_C$.
\vspace{-3mm}
\subsubsection{Trailer Decoder} \label{sec:decoder}
Given the output of the context encoder, we utilize a trailer decoder $\calD_\calT$ to autoregressively generate shot embeddings, which are subsequently matched back to the movie sequence to compose the trailer. At training time, the trailer decoder inputs the learned context $C$ and a positionally encoded but right-shifted trailer sequence as shown in~\Fref{fig:overview} (a). The decoding is done in a sequential manner that $\calD_\calT$ attends to the context $C$ and a trailer sequence up to the current position $\{\mathrm{\texttt{SOS}}, v_1,\ldots, v_{j-1}\}$ in order to output the next trailer embedding $\hat{v}_j$ (\Eref{eqn: decode}). This enables the network to efficiently capture sequential dependencies and positional information of shots, facilitating the learning of shot composition for generating a trailer sequence. The next trailer shot generation scheme in our work is equivalent to next-word prediction in language modeling~\cite{vaswani2017attention}. We use a stack of Transformer~\cite{vaswani2017attention} decoder layers for $\calD_\calT$. At inference time, as ground truth trailer sequence is not available, the trailer decoder starts with $(C, \{\texttt{SOS}\})$ and autoregressively decodes a trailer sequence using the previously generated trailer shot embeddings until decoding the \texttt{EOS} token. The decoding process is formulated as follows:
\begin{equation}
    \hat{v}_j = \calD_\calT (C, \{\texttt{SOS}, v_1, \ldots, v_{j-1}\}),~~~~~1\le j\le m+1.
    \label{eqn: decode}
\end{equation}
Lastly, we adopt a Greedy Search strategy utilizing Nearest Neighbor retrieval. This method involves comparing each embedding of the decoded shots against all shot representations from the movie. The shot with the highest feature similarity is then selected for assembling the output trailer.

\vspace{-2mm}
\subsubsection{Training Losses} We train our network by optimizing the embedding distance between the predicted trailer sequence $\{\hat{v}_1,\ldots, \hat{v}_m, \hat{v}_{m+1}\}$ and the ground truth sequence $\{v_1, \ldots, v_m, \texttt{EOS}\}$. To ensure that the trailer decoder outputs the correct shot at each step, we minimize the reconstruction loss between each predicted trailer embedding and the corresponding ground truth embedding as shown in \Eref{eqn: rec}. As mentioned earlier, the order in which the shots are decoded is crucial for trailer generation. To integrate this concept into the proposed model, we introduce a sequence-based loss, minimizing the Kullback-Leibler (KL) divergence loss between the predicted trailer sequence distribution and the ground truth distribution, as shown in \Eref{eqn: kl}.
\begin{equation}
    \calL_\mathrm{rec} = \sum_{j=1}^{m+1}|\hat{v}_j - v_j|^{2},
    \label{eqn: rec}
\end{equation}
\vspace{-4mm}
\begin{equation}
    \calL_\mathrm{KL} = \sum_j \texttt{softmax}(v_j) \cdot \log\Big(\frac{\texttt{softmax}(v_j)}{\texttt{softmax}(\hat{v}_j)}\Big).
    \label{eqn: kl}
\end{equation}
The total training loss is defined as the sum of the trailerness encoding loss in \Eref{eqn: loss_trailerness}, the feature reconstruction loss in \Eref{eqn: rec}, and the sequence-based loss \Eref{eqn: kl}.
\begin{equation}
    \calL_{\mathrm{total}} = \calL_t + \calL_\mathrm{rec} +  \calL_\mathrm{KL}.
\end{equation}

\subsection{Datasets}
Movies exhibit a large diversity encompassing genre, style, narrative structure, theme, artistic choices, and many more. We argue that exposing our model to a large and comprehensive collection of movies is crucial for building a generalizable trailer generation pipeline. Following this motivation, we collect a training set of 23,304 movies and trailers spanning 28 genres. These movies cover over 120 years of cinematic history, albeit with a natural bias towards more recent films, reflecting the industry's continuous growth and increased production rates. The training set is paired with a validation split containing 300 movie-trailer pairs used for hyperparameter selection. 

Furthermore, we introduce new benchmarks for trailer generation, building upon two mainstream movie datasets: MAD~\cite{soldan2022mad} and MovieNet~\cite{huang2020movienet}. 
To align these datasets with our task, we enhance them by including their respective trailers obtained from IMDb~\cite{IMDb}. Notably, not all movies in these datasets have trailers available on their IMDb pages. Consequently, we obtained 602 out of 650 movie-trailer pairs for MAD and 989 out of 1100 for MovieNet. We also ensure no overlap between the test sets and the training/validation splits.

\subsection{Evaluation Metrics} \label{sec:metrics}
We evaluate the trailer generation performance by taking both shot selection and sequencing into account. We follow previous works~\cite{hesham2018smart,wang2020learning} and employ Precision, Recall, and F1-score metrics to measure the accuracy of a model in predicting trailer shots from an input movie sequence. We also measure the correctness of the predicted trailer sequence using Levenshtein distance (LD) and Sequence length difference (SLD) metrics. Levenshtein  distance (also known a edit distance)~\cite{levenshtein1966binary} is defined as the minimum number of edits (insertions, deletions, or substitutions) required to transform the decoded trailer sequence into the ground truth sequence. Sequence length difference measures the absolute difference in the number of shots between the predicted trailer sequence and the ground truth sequence.

\section{Experiment}\label{sec:experiments}

\begin{table*}[!t]
\begin{center}
    \caption{\textbf{Experimental comparison with different baselines.}}
    \vspace{-0.3cm}
    \mytabular{0.8}{
    \begin{tabular}{lcccccccccc}
    \toprule
    &  \multicolumn{5}{c}{MAD~\cite{soldan2022mad}} 
    & \multicolumn{5}{c}{MovieNet~\cite{huang2020movienet}} \\ 
    
    \cmidrule(lr){2-6} \cmidrule(lr){7-11} 
    
    & Precision $\uparrow$ & Recall $\uparrow$ & F1-score $\uparrow$ & LD $\downarrow$ & SLD $\downarrow$ 
    & Precision $\uparrow$ & Recall $\uparrow$ & F1-score $\uparrow$ & LD $\downarrow$ & SLD $\downarrow$\\ 
    
    \midrule
    
    Random & 5.32 & 6.05 & 5.65 & - & - 
           & 4.96 & 5.67 & 5.28 & - & -\\
    
    CCANet~\cite{wang2020learning} 
    & 32.15 & 30.76 & 31.63 & 81.25 & - 
    & 30.46 & 29.29 & 29.58 & 90.18 &-\\
    
    CLIP-It~\cite{narasimhan2021clip} 
    & 40.29 & 43.05 & 41.73 &  95.58 & 47.10 
    & 38.19 & 40.28 & 39.34 & 103.64 & 51.30\\
    
    \textbf{TGT (ours) }
    & \textbf{55.30} & \textbf{49.92} & \textbf{52.38} & \textbf{21.18} & \textbf{10.78} 
    & \textbf{49.74} & \textbf{44.32} & \textbf{46.77} & \textbf{24.66} & \textbf{13.28}\\
    
    \bottomrule
    \end{tabular}
    }
    \label{tab:baselines_comparison}
    \vspace{-0.5cm}
\end{center}
\end{table*}

\begin{figure*}[!t]
\begin{center}
\setlength{\tabcolsep}{0.5pt}
\renewcommand{\arraystretch}{0.5}
\resizebox{1.0\linewidth}{!}{%
\begin{tabular}{ccccccc}
\includegraphics[width=0.1\linewidth]{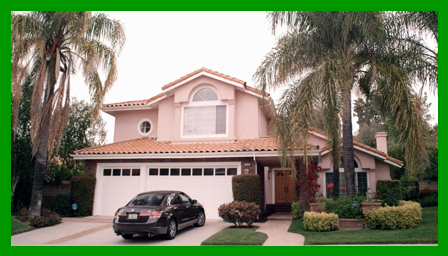}&
\includegraphics[width=0.1\linewidth]{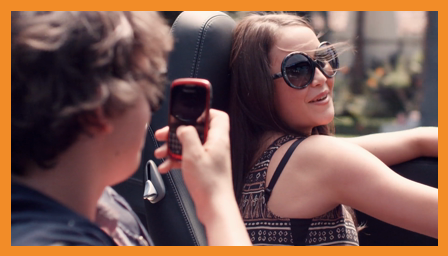}&
\includegraphics[width=0.1\linewidth]{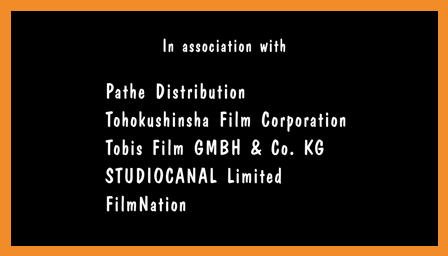}&
\includegraphics[width=0.1\linewidth]{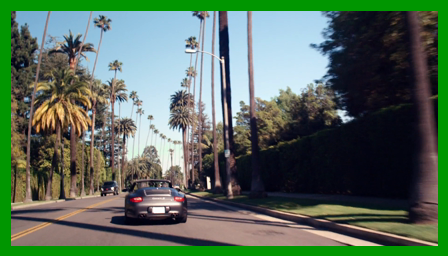}&
\includegraphics[width=0.1\linewidth]{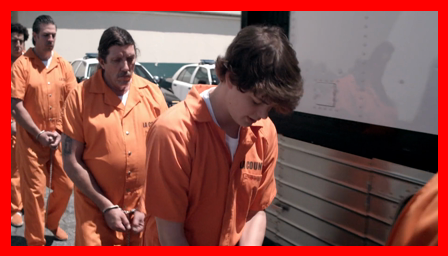}&
\includegraphics[width=0.1\linewidth]{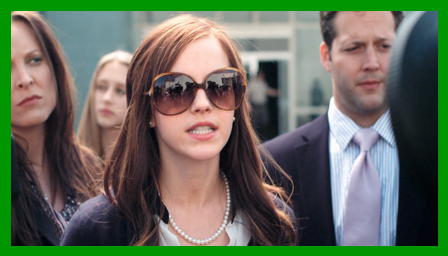}& 
\raisebox{0.8\normalbaselineskip}{\rotatebox[origin=c]{90}{\tiny{Ours}}}\\

\includegraphics[width=0.1\linewidth]{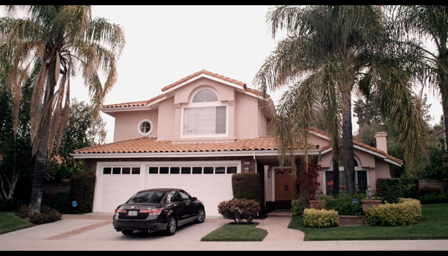}&
\includegraphics[width=0.1\linewidth]{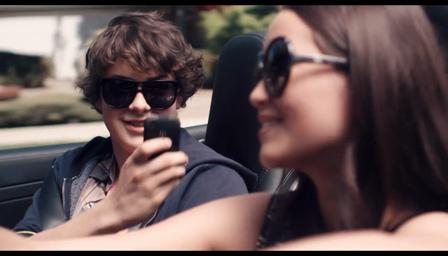}&
\includegraphics[width=0.1\linewidth]{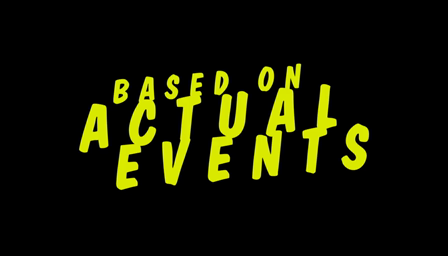}&
\includegraphics[width=0.1\linewidth]{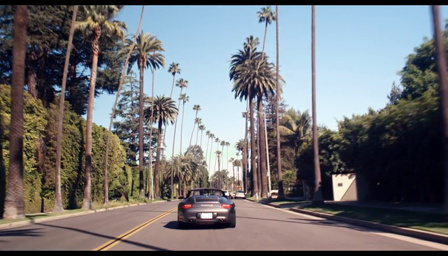}&
\includegraphics[width=0.1\linewidth]{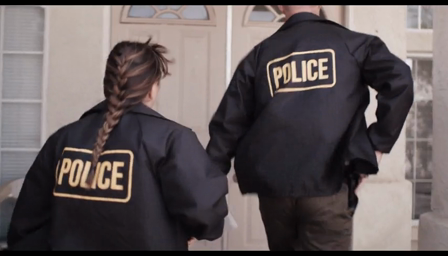}&
\includegraphics[width=0.1\linewidth]{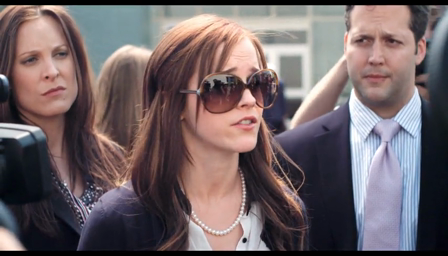} & 
\raisebox{0.8\normalbaselineskip}{\rotatebox[origin=c]{90}{\tiny{GT}}}
\end{tabular}}

\end{center}
\vspace{-0.5cm}
\caption{\textbf{Subjective quality of our trailer generation.}
We compare a movie trailer against the one produced by our TGT method. We highlight in green the correctly selected shots, in orange shots that are visually similar, and in red mismatched shots. 
}
\label{fig: qual_compare}
\vspace{-0.5cm}
\end{figure*}

\paragraph{Implementation Details} 
We use the TransNet-V2~\cite{souvcek2020transnet} model for preprocessing movies and trailers into sequences of shots. Additionally, we employ the pretrained CLIP ViT H/14~\cite{radford2021learning} for shot feature extraction. The trailerness encoder $\calE_\calT$ contains a single Transformer~\cite{vaswani2017attention} encoder layer. The context encoder $\calE_\calC$ consists of four encoder layers. Five decoder layers are used for the trailer decoder $\calD_\calT$. Each encoder and decoder layer has a hidden dimension of 1024, 8 attention heads, and a feed-forward dimension of 2048. \texttt{SOS} and \texttt{EOS} tokens are initialized by creating an embedding vector of size 1024, sampled from a Normal distribution. AdamW~\cite{loshchilov2017decoupled} optimizer with an initial learning rate of $1e-4$ and a cosine warm-up scheduler is used during training. Our model is trained for 200 epochs with a mini-batch size of 8, on a single NVIDIA A100 GPU~\footnote{Training experiments done at KAUST.}. 

\subsection{Trailer Generation Performance}
In this section, we evaluate the performance of our approach by comparing it with three baseline methods: Random, CCANet~\cite{wang2020learning} and CLIP-It~\cite{narasimhan2021clip}, on the MAD~\cite{soldan2022mad} and MovieNet~\cite{huang2020movienet} benchmarks. To establish a Random baseline, we randomly select $m$ shots from each movie, where $m$ represents the total number of shots present in the corresponding ground truth trailer. 
As shown in~\Tref{tab:baselines_comparison}, a baseline that randomly selects movie shots for trailer generation exhibits significantly poor performance across all metrics.
\vspace{-3mm}
\paragraph{Comparison to Trailer Generation SoTA}
CCANet~\cite{wang2020learning} employs a co-attention module to rank a set of movie shots in a contrastive manner, where shots with higher rankings are considered to be trailer moments. To assess the performance of CCANet for trailer generation, we select the top $m$ ranked shots from a given movie and utilize the resulting sequence for evaluation purposes. As evident from  \Tref{tab:baselines_comparison}, our approach outperforms CCANet by a large margin. For instance, on the F1-score metric, TGT achieves an average accuracy of 52.38\% and 46.77\% on MAD~\cite{soldan2022mad} and MovieNet~\cite{huang2020movienet} datasets, respectively. In comparison, CCANet could only achieve 31.63\% and 29.58\%. This is mainly because CCANet has a limited attention span, focusing on two shots at a time, which hinders accurate rankings due to the abundance of negative samples in full-length movies. In contrast, our approach generates trailers by simultaneously considering the entire movie sequence. It can also be inferred from \Tref{tab:baselines_comparison} that the trailer sequence decoded by TGT shows much greater similarity to the ground truth sequence ($\sim$21 edits on LD metric on MAD dataset) compared to CCANet ($\sim$81 edits) even though the LD metric favors CCANet due to the assumption of an equal sequence length to the ground truth. This is an expected observation since CCANet~\cite{wang2020learning} mainly targets trailer moment detection and does not take shot sequencing into account. 

\vspace{-3mm}
\paragraph{Comparison to Video Summarization SoTA}
We conduct further comparison by adapting a state-of-the-art video summarization model, CLIP-It~\cite{narasimhan2021clip}, for trailer generation. CLIP-It utilizes a vanilla Transformer~\cite{vaswani2017attention} architecture, where the input sequence is passed through both the encoder and decoder. Additionally, a binary classifier is employed to identify summary-worthy inputs in the sequence.
In a parallel fashion, we classify each shot in a movie as either belonging to the trailer or not. The shots classified as trailer shots are subsequently merged to construct the final trailer. The results presented in \Tref{tab:baselines_comparison} suggest that the CLIP-It~\cite{narasimhan2021clip} based method generally achieves better performance compared to CCANet~\cite{wang2020learning}. However, it is worth highlighting that our approach notably outperforms CLIP-It. For example, on the Recall metric, TGT outperforms CLIP-It by 15.95\% and 10.03\% on MAD~\cite{soldan2022mad} and MovieNet~\cite{huang2020movienet} benchmarks, respectively. This is likely because CLIP-It~\cite{narasimhan2021clip} primarily focuses on shot-level classification and may not fully capture the temporal correlations between shots necessary for creating a compelling trailer. In contrast, our autoregressive approach enables TGT to attend to the contextual dependencies between shots, aligning with the movie's narrative and dramatic flow to generate a coherent and structured trailer. In terms of the predicted trailer length, it is evident from \Tref{tab:baselines_comparison} that CLIP-It deviates by $\sim$51 shots on the MovieNet dataset, whereas TGT only deviates by $\sim$11 shots on average.
\vspace{-3mm}
\paragraph{Qualitative Results}
In \Fref{fig: qual_compare}, we qualitatively compare a trailer decoded by our approach with its corresponding ground truth sequence. For visualization purposes, we select a subsequence from the original trailer.
The shots enclosed in \textcolor{green}{green} boxes represent accurate predictions, while those in \textcolor{orange}{orange} boxes indicate resembling predictions. The shots in \textcolor{red}{red} boxes correspond to failed predictions. The results in \Fref{fig: qual_compare} indicate that the proposed method outputs a reasonable trailer sequence from a given movie unseen during training. For instance, TGT generates a third shot that closely resembles the corresponding insert shot (black frame with text) in the ground truth, even though the movie sequence itself does not contain that specific shot. 
\vspace{-1mm}
\subsection{Ablation Studies} \label{sec:ablations}

\paragraph{Network Components}
In \Tref{tab:network_ablation}, we present ablation experiments on the different network components used in the proposed framework. First, we study the importance of trailerness encoding by training our network without $\calE_\calT$, \ie~we directly feed the positionally encoded movie sequence into the context encoder. As can be inferred from \Tref{tab:network_ablation}, TGT trained without $\calE_\calT$ still gives a competitive performance across all metrics. This is intuitive since the context encoder has to implicitly capture the trailerness of each shot as it is trained to learn a contextualized representation for the trailer decoder. However, explicitly embedding relative trailerness scores (predicted from  $\calE_\calT$) into the movie sequence has resulted in a better performance. For instance, on the F1-score metric, trailerness encoding results in a 4.35\% and 4.53\% performance improvement on MAD~\cite{soldan2022mad} and MovieNet~\cite{huang2020movienet}. 

\begin{table}[!t]
\begin{center}
    \caption{\textbf{Ablation experiment on network components.} $\calE_\calT$: trailerness encoder; $\calE_\calC$: context encoder. It shows both encoders contribute to  TGT performance, and $\calE_\calC$ is especially important.} 
    \vspace{-0.1cm}
    \mytabular{0.75}{
    \begin{tabular}{llcccccccc}
    
    \toprule
    
    && Precision $\uparrow$ & Recall $\uparrow$ & F1-score $\uparrow$ & LD $\downarrow$ \\ 
    
    \midrule

    \multirow{4}{*}{\rotatebox[origin=c]{90}{MAD}}
    & w/o $\calE_\calT$ 
    & 53.23 & 47.64 & 50.20 & 22.72 \\
    
    & w/o $\calE_\calC$ 
    & 43.36 & 36.07 & 39.23 & 25.70 \\
   
    & w/o $\calE_\calT + \calE_\calC$ 
    & 42.36 & 36.15 & 38.89 & 25.75 \\
    
    & {w/ $\calE_\calT + \calE_\calC$ (\textbf{TGT})}
    & \textbf{55.30} & \textbf{49.92} & \textbf{52.38} & \textbf{21.18} \\
       
    \midrule

     \multirow{4}{*}{\rotatebox[origin=c]{90}{MovieNet}}
     & w/o $\calE_\calT$ 
     & 48.38 & 41.80 & 44.75 & 25.62\\
     
     & w/o $\calE_\calC$ 
     & 40.08 & 33.41 & 36.32 & 27.09\\
     
     & w/o $\calE_\calT + \calE_\calC$ 
     & 39.01 & 33.01 & 35.64 & 27.86\\
     
     & {w/ $\calE_\calT + \calE_\calC$ (\textbf{TGT})}
     & \textbf{49.74} & \textbf{44.32} & \textbf{46.77} & \textbf{24.66}\\
    
    \bottomrule
    
    \end{tabular}
    }
    \label{tab:network_ablation}
    \vspace{-0.7cm}
\end{center}
\end{table}

To investigate the benefit of learning contextualized movie representation for trailer generation, we train our network without the context encoder $\calE_\calC$. To do so, we use the trailerness encoded movie sequence as a context and feed it into the trailer decoder. As shown in \Tref{tab:network_ablation}, training TGT without $\calE_\calC$ results in a notably worse performance. For example, on the Recall metric, TGT without $\calE_\calC$ could only achieve a top-1 accuracy of 36.07\% and 33.41\% on MAD~\cite{soldan2022mad} and MovieNet~\cite{huang2020movienet} datasets, respectively. In comparison, TGT with $\calE_\calC$ gives an accuracy of 49.92\% and 44.32\%. It can be observed from \Tref{tab:network_ablation} that a similar performance drop happens when using a decoder-only architecture, \ie~the positionally encoded movie sequence is fed into the trailer decoder $\calD_\calT$ (without $\calE_\calT + \calE_\calC$). These results demonstrate the challenge of directly generating a trailer sequence from a movie sequence and highlight the importance of obtaining a contextualized representation of the movie for effective trailer decoding.

\begin{table}[!t]
\begin{center}
    \caption{\textbf{Ablation experiment on loss functions.} $\calL_\mathrm{rec}$: reconstruction loss of trailer embeddings; $\calL_t$: trailerness loss; $\calL_\mathrm{KL}$: KL divergence loss of trailer sequence distribution. It shows that both $\calL_t$ and $\calL_\mathrm{KL}$ are beneficial, and using them together with $\calL_\mathrm{rec}$ leads to the highest performance.   }
    \vspace{-0.1cm}
    \mytabular{0.73}{
    \begin{tabular}{llcccc}
    
    \toprule
    
    && Precision $\uparrow$ & Recall $\uparrow$ & F1-score $\uparrow$ & LD $\downarrow$\\ 
    
    \midrule

    \multirow{4}{*}{\rotatebox[origin=c]{90}{MAD}}
    & $\calL_\mathrm{rec}$ 
    & 49.74 & 43.61 & 46.38 & 26.81 \\
    
    & $\calL_\mathrm{rec} + \calL_t$ 
    & 52.35 & 46.06 & 48.90 & 25.84 \\
    
    & $\calL_\mathrm{rec} + \calL_\mathrm{KL}$ 
    & 53.86 & 48.63 & 51.03 & 21.99 \\
    
    & $\calL_\mathrm{rec} + \calL_t + \calL_\mathrm{KL}$ (\textbf{TGT})
    & \textbf{55.30} & \textbf{49.92} & \textbf{52.38} & \textbf{21.18} \\

    \midrule

    \multirow{4}{*}{\rotatebox[origin=c]{90}{MovieNet}}
    & $\calL_\mathrm{rec}$ 
    & 45.40 & 39.32 & 42.04 & 31.84\\

    & $\calL_\mathrm{rec} + \calL_t$ 
    & 46.52 & 40.15 & 42.99 & 30.91\\
    
    & $\calL_\mathrm{rec} + \calL_\mathrm{KL}$ 
    & 48.48 & 43.19 & 45.60 & 25.19\\

    & $\calL_\mathrm{rec} + \calL_t + \calL_\mathrm{KL}$ (\textbf{TGT})
    & \textbf{49.74} & \textbf{44.32} & \textbf{46.77} & \textbf{24.66}\\
    
    \bottomrule
    
    \end{tabular}
    }
    \label{tab:loss_ablation}
    \vspace{-0.7cm}
\end{center}
\end{table}

\vspace{-2mm}
\paragraph{Loss Functions} 
Here, we examine the contribution of various loss functions used to train our network. First, we train TGT using the reconstruction loss $\calL_\mathrm{rec}$ in \Eref{eqn: rec} as the only loss function. As can be seen from \Tref{tab:loss_ablation}, $\calL_\mathrm{rec}$ is a strong enough constraint to train a competitive baseline model. We also investigate the benefit of optimizing the predicted trailerness scores with ground truth scores. It can be inferred from \Tref{tab:loss_ablation} that TGT trained with trailerness loss $\calL_t$ in \Eref{eqn: loss_trailerness} performs consistently better compared to TGT trained without $\calL_t$. For instance, on the Precision metric, a network trained with $\calL_\mathrm{rec} + \calL_t$ outperforms a network trained with $\calL_\mathrm{rec}$ by 5.25\% and 2.46\% on MAD~\cite{soldan2022mad} and MovieNet~\cite{huang2020movienet} datasets, respectively.

We investigate the impact of minimizing the KL divergence between the predicted trailer sequence and the ground truth sequence as defined in \Eref{eqn: kl}. The results in \Tref{tab:loss_ablation} show that training with $\calL_\mathrm{KL}$ results in a considerable performance gain. Particularly, on the Levenshtein distance (LD) metric, a trailer decoded by TGT trained with $\calL_\mathrm{KL}$ requires an average of 4.83 and 6.65 fewer edit steps (to change it to the ground truth sequence) on MAD~\cite{soldan2022mad} and MovieNet~\cite{huang2020movienet} datasets, respectively, compared to TGT trained without $\calL_\mathrm{KL}$. These results emphasize the significance of the KL-divergence loss $\calL_\mathrm{KL}$ in ensuring that the trailer sequence generated by TGT closely matches the ground truth sequence. As can be inferred from \Tref{tab:loss_ablation}, the combination of all training losses ($\calL_\mathrm{rec} + \calL_t + \calL_\mathrm{KL}$) yields the best network performance.

\subsection{Experimental Analyses}
\paragraph{Text-Controlled Trailer Generation}
Thus far, we have formulated movie trailer generation as a video-to-video problem. Here, we experiment with incorporating language into TGT to guide trailer decoding. For this purpose, we obtain the plot summary of each movie from IMDb~\cite{IMDb} and encode the text using a pretrained RoBERTa~\cite{liu2019roberta} model. We explore two types of network variants for text-conditioned trailer generation. First, we simply concatenate the encoded text with the output of the context encoder as a control to enrich the context pool of TGT, and the resulting representation will be fed into the trailer decoder (referred to as ``encoded text'' in \Tref{tab:analysis_controlled_generation}). Second, we experiment with reasoning over the encoded text before feeding it to the trailer decoder. We use a single Transformer~\cite{vaswani2017attention} encoder layer to obtain a contextualized text representation (referred to as ``contextualized text'' in \Tref{tab:analysis_controlled_generation}). As evident from \Tref{tab:analysis_controlled_generation}, ``encoded text'' leads to a significant improvement in trailer decoding performance across all metrics. Using ``contextualized text'' on top of pretrained RoBERTa~\cite{liu2019roberta} further improves trailer generation performance. This finding is consistent with our expectation that additional context enhances generation quality. More importantly, as a plot summary typically includes major plot points such as the setting, characters, and conflicts, it enables the trailer decoder to establish a more structured trailer decoding rule by attending to the relevant movie shots.  

\begin{table}[!t]
\renewcommand{\arraystretch}{1.1}
\begin{center}
    \vspace{-0.2cm}
    \caption{\textbf{Analysis on text-controlled trailer generation.} ``Encoded text'' leads to a significant improvement across all metrics. Using ``contextualized text'' further improves the performance.}
    \vspace{-0.1cm}
    \mytabular{0.75}{
    \begin{tabular}{llcccc}
    
    \toprule
    
    && Precision $\uparrow$ & Recall $\uparrow$ & F1-score $\uparrow$ & LD $\downarrow$ \\
    
    \midrule

    \multirow{3}{*}{\rotatebox[origin=c]{90}{MAD}} 
    &  TGT 
    & 55.30 & 49.92 & 52.38 & 21.18  \\
    
    & + Encoded text 
    & 59.27 & 54.68 & 56.81 & 19.34 \\
    
    & + Contextualized text 
    & \textbf{60.93} & \textbf{56.44} & \textbf{58.52} & \textbf{18.40}\\ 

    \midrule

    \multirow{3}{*}{\rotatebox[origin=c]{90}{MovieNet}} 
    & TGT 
    & 49.74 & 44.32 & 46.77 & 24.66 \\
    
    & + Encoded text 
    & 53.69 & 49.05 & 51.19 & 23.59\\
    
    & + Contextualized text 
    & \textbf{56.13} & \textbf{51.65} & \textbf{53.72} & \textbf{21.81}\\ 
    
    \bottomrule
    \end{tabular}
    }
    \label{tab:analysis_controlled_generation}
    \vspace{-0.5cm}
\end{center}
\end{table}

\begin{table}[!t]
\renewcommand{\arraystretch}{1.1}
\begin{center}
    \caption{\textbf{Analysis on trailer shot selection.} Relaxing the metric enhances performance, indicating the potential of TGT to be used in collaboration with a human editor.}
    \vspace{-0.1cm}
    \mytabular{0.85}{
    \begin{tabular}{llcccc}
    
    \toprule
    
    & & Precision $\uparrow$ & Recall $\uparrow$ & F1-score $\uparrow$ & LD $\downarrow$  \\
    
    \midrule
    
    \multirow{3}{*}{\rotatebox[origin=c]{90}{MAD}} 
    & Top-1  & 55.30 & 49.92 & 52.38 & 21.18 \\
    & Top-5  & 65.31 & 60.57 & 62.76 & 15.51 \\
    & Top-10 & \textbf{69.02} & \textbf{64.85} & \textbf{66.79} & \textbf{13.05} \\

    \midrule

    \multirow{3}{*}{\rotatebox[origin=c]{90}{MovieNet}} 
    & Top-1  & 49.74 & 44.32 & 46.77 & 24.66 \\
    & Top-5  & 58.58 & 54.05 & 56.12 & 18.86 \\
    & Top-10 & \textbf{61.84} & \textbf{57.74} & \textbf{59.63} & \textbf{16.23} \\ 
    
    \bottomrule
    
    \end{tabular}
    }
    \label{tab:analysis_shot_selection}
    \vspace{-0.7cm}
\end{center}
\end{table}

\paragraph{Shot Selection}
One of the key challenges of generating a trailer from a movie is shot selection. At each step of decoding, there could be multiple movie shot candidates that are semantically similar to the target trailer shot. In \Tref{tab:analysis_shot_selection}, we evaluate our model's performance by analyzing the quality of the shots decoded at each step, considering the top-5 and top-10 matching shots. As anticipated, expanding the pool of shot candidates leads to enhanced model performance across all evaluated metrics. For instance, on MAD~\cite{soldan2022mad} benchmark, TGT achieves a top-10 precision of 69.02\% and the resulting trailer is only $\sim$13 edit steps away from the ground truth trailer. This is particularly noteworthy as it showcases the potential of our model to serve as a shot recommender during the editing process, where a human editor collaborates with the model to make informed decisions. 

\vspace{-0.3cm}\paragraph{Scaling Effect}
In \Tref{tab:analysis_scale}, we present the results of a scaling experiment conducted to evaluate the performance of our approach using different training data sizes, specifically 10\%, 50\%, and 100\% of the available data. As can be inferred from the table, model performance directly correlates with the size of the training data. This observation is intuitive, as a larger dataset enables the model to encompass a broader range of patterns and variations found in movie and trailer sequences. 

\begin{table}[!t]
\renewcommand{\arraystretch}{1.1}
\vspace{-0.2cm}
\begin{center}
    \caption{\textbf{Analysis on the training data size.} 
Performance drops significantly when using only 10\% of the data, while using 50\% of the data yields decent performance compared to the full dataset.}
\vspace{-0.1cm}
    \mytabular{0.75}{
    \begin{tabular}{llcccc}
    
    \toprule

    && Precision $\uparrow$ & Recall $\uparrow$ & F1-score $\uparrow$ & LD $\downarrow$ \\ 
    
    \midrule

    \multirow{3}{*}{\rotatebox[origin=c]{90}{MAD}} 
    & 10 \% 
    & 35.23 & 33.25 & 34.11 & 27.34  \\
    
    & 50 \% 
    & 50.30 & 47.71 & 48.80 &24.32 \\
    
    & 100 \% 
    & \textbf{55.30} & \textbf{49.92} & \textbf{52.38} & \textbf{21.18} \\

    \midrule

    \multirow{3}{*}{\rotatebox[origin=c]{90}{MovieNet}} 
    & 10 \% 
    & 28.38 & 26.81 & 27.46 & 30.84 \\
    
    & 50 \% 
    & 44.71 & 41.61 & 42.96 & 27.43 \\
    
    & 100 \% 
    & \textbf{49.74} & \textbf{44.32} & \textbf{46.77} & \textbf{24.66} \\
    
    \bottomrule
    \end{tabular}
    }
    \label{tab:analysis_scale}
\end{center}
\vspace{-0.7cm}
\end{table}

\subsection{Discussion and Limitations}
\vspace{-0.5mm}
The proposed TGT method streamlines the trailer creation process by enabling efficient selection and ordering of shots. However, the current method does not incorporate dialogue and sound modeling, which are crucial for fine editing. This limitation could be addressed in future work by incorporating these additional factors into the TGT model. Despite this limitation, the TGT method can arguably provide significant time-saving to editors by automating the initial steps of shot selection and ordering. This shift in workload empowers editors to focus on the artistic aspects of trailer creation, such as refining cuts, durations, and injecting subtle audio elements to enhance the trailer further. Therefore, the TGT method, while limited in scope, has the potential to significantly improve the efficiency and creativity of trailer creation.

\vspace{-1mm}
\section{Conclusion}\label{sec:conclusions}
\vspace{-0.5mm}
This work presents a novel approach to automatic trailer generation. Our TGT models the task as a machine translation problem and uses an effective encode-decoder architecture to generate plausible trailers. We present two newly constructed benchmarks and show that TGT outperforms state-of-the-art approaches. We believe this study has the potential to advance video summarization and promotional content creation across various domains.
\vspace{-2mm}
\paragraph{Acknowledgement} This work was supported by the King Abdullah University of Science and Technology (KAUST) Office of Sponsored Research through the Visual Computing Center (VCC) funding, as well as the SDAIA-KAUST Center of Excellence in Data Science and Artificial Intelligence (SDAIA-KAUST AI).
{
    \small
    \bibliographystyle{ieeenat_fullname}
    \bibliography{main}

\begin{thebibliography}{44}
\providecommand{\natexlab}[1]{#1}
\providecommand{\url}[1]{\texttt{#1}}
\expandafter\ifx\csname urlstyle\endcsname\relax
  \providecommand{\doi}[1]{doi: #1}\else
  \providecommand{\doi}{doi: \begingroup \urlstyle{rm}\Url}\fi

\bibitem[IMD()]{IMDb}
Imdb.
\newblock \url{https://www.imdb.com/}.
\newblock Accessed: March 15, 2023.

\bibitem[Badamdorj et~al.(2022)Badamdorj, Rochan, Wang, and Cheng]{badamdorj2022contrastive}
Taivanbat Badamdorj, Mrigank Rochan, Yang Wang, and Li Cheng.
\newblock Contrastive learning for unsupervised video highlight detection.
\newblock In \emph{Proceedings of the IEEE/CVF Conference on Computer Vision and Pattern Recognition}, pages 14042--14052, 2022.

\bibitem[Bahdanau et~al.(2014)Bahdanau, Cho, and Bengio]{bahdanau2014neural}
Dzmitry Bahdanau, Kyunghyun Cho, and Yoshua Bengio.
\newblock Neural machine translation by jointly learning to align and translate.
\newblock \emph{arXiv preprint arXiv:1409.0473}, 2014.

\bibitem[Chen et~al.(2004)Chen, Kuo, Chu, and Wu]{chen2004action}
Hsuan-Wei Chen, Jin-Hau Kuo, Wei-Ta Chu, and Ja-Ling Wu.
\newblock Action movies segmentation and summarization based on tempo analysis.
\newblock In \emph{Proceedings of the 6th ACM SIGMM international workshop on Multimedia information retrieval}, pages 251--258, 2004.

\bibitem[Fajtl et~al.(2019)Fajtl, Sokeh, Argyriou, Monekosso, and Remagnino]{fajtl2019summarizing}
Jiri Fajtl, Hajar~Sadeghi Sokeh, Vasileios Argyriou, Dorothy Monekosso, and Paolo Remagnino.
\newblock Summarizing videos with attention.
\newblock In \emph{Computer Vision--ACCV 2018 Workshops: 14th Asian Conference on Computer Vision, Perth, Australia, December 2--6, 2018, Revised Selected Papers 14}, pages 39--54. Springer, 2019.

\bibitem[Gaikwad et~al.(2021)Gaikwad, Sontakke, Patwardhan, Pedanekar, and Karande]{Gaikwad_2021_ICCV}
Bhagyashree Gaikwad, Ankita Sontakke, Manasi Patwardhan, Niranjan Pedanekar, and Shirish Karande.
\newblock Plots to previews: Towards automatic movie preview retrieval using publicly available meta-data.
\newblock In \emph{Proceedings of the IEEE/CVF International Conference on Computer Vision (ICCV) Workshops}, pages 3205--3214, 2021.

\bibitem[Gan et~al.(2023)Gan, Shu, Qiao, Wu, Chen, Li, and Ren]{gan2023collaborative}
Bei Gan, Xiujun Shu, Ruizhi Qiao, Haoqian Wu, Keyu Chen, Hanjun Li, and Bo Ren.
\newblock Collaborative noisy label cleaner: Learning scene-aware trailers for multi-modal highlight detection in movies.
\newblock \emph{arXiv preprint arXiv:2303.14768}, 2023.

\bibitem[Gygli et~al.(2014)Gygli, Grabner, Riemenschneider, and Van~Gool]{gygli2014creating}
Michael Gygli, Helmut Grabner, Hayko Riemenschneider, and Luc Van~Gool.
\newblock Creating summaries from user videos.
\newblock In \emph{Computer Vision--ECCV 2014: 13th European Conference, Zurich, Switzerland, September 6-12, 2014, Proceedings, Part VII 13}, pages 505--520. Springer, 2014.

\bibitem[Hesham et~al.(2018)Hesham, Hani, Fouad, and Amer]{hesham2018smart}
Mohammad Hesham, Bishoy Hani, Nour Fouad, and Eslam Amer.
\newblock Smart trailer: Automatic generation of movie trailer using only subtitles.
\newblock In \emph{2018 First International Workshop on Deep and Representation Learning (IWDRL)}, pages 26--30. IEEE, 2018.

\bibitem[Huang et~al.(2020)Huang, Xiong, Rao, Wang, and Lin]{huang2020movienet}
Qingqiu Huang, Yu Xiong, Anyi Rao, Jiaze Wang, and Dahua Lin.
\newblock Movienet: A holistic dataset for movie understanding.
\newblock In \emph{Computer Vision--ECCV 2020: 16th European Conference, Glasgow, UK, August 23--28, 2020, Proceedings, Part IV 16}, pages 709--727. Springer, 2020.

\bibitem[Irie et~al.(2010)Irie, Satou, Kojima, Yamasaki, and Aizawa]{irie2010automatic}
Go Irie, Takashi Satou, Akira Kojima, Toshihiko Yamasaki, and Kiyoharu Aizawa.
\newblock Automatic trailer generation.
\newblock In \emph{Proceedings of the 18th ACM international conference on Multimedia}, pages 839--842, 2010.

\bibitem[Kawai et~al.(2007)Kawai, Sumiyoshi, and Yagi]{kawai2007automated}
Yoshihiko Kawai, Hideki Sumiyoshi, and Nobuyuki Yagi.
\newblock Automated production of tv program trailer using electronic program guide.
\newblock In \emph{Proceedings of the 6th ACM international conference on Image and video retrieval}, pages 49--56, 2007.

\bibitem[Khosla et~al.(2013)Khosla, Hamid, Lin, and Sundaresan]{khosla2013large}
Aditya Khosla, Raffay Hamid, Chih-Jen Lin, and Neel Sundaresan.
\newblock Large-scale video summarization using web-image priors.
\newblock In \emph{Proceedings of the IEEE conference on computer vision and pattern recognition}, pages 2698--2705, 2013.

\bibitem[King et~al.(2021)King, Zavesky, and Gonzales]{king2021user}
Allyson King, Eric Zavesky, and Michael~J Gonzales.
\newblock User preferences for automated curation of snackable content.
\newblock In \emph{26th International Conference on Intelligent User Interfaces}, pages 270--274, 2021.

\bibitem[Levenshtein et~al.(1966)]{levenshtein1966binary}
Vladimir~I Levenshtein et~al.
\newblock Binary codes capable of correcting deletions, insertions, and reversals.
\newblock In \emph{Soviet physics doklady}, pages 707--710. Soviet Union, 1966.

\bibitem[Li et~al.(2018)Li, Wang, Yang, and Gong]{li2018local}
Yandong Li, Liqiang Wang, Tianbao Yang, and Boqing Gong.
\newblock How local is the local diversity? reinforcing sequential determinantal point processes with dynamic ground sets for supervised video summarization.
\newblock In \emph{Proceedings of the European Conference on Computer Vision (ECCV)}, pages 151--167, 2018.

\bibitem[Liu and Jiang(2015)]{liu2015semi}
Xingchen Liu and Jianming Jiang.
\newblock Semi-supervised learning towards computerized generation of movie trailers.
\newblock In \emph{2015 IEEE International Conference on Systems, Man, and Cybernetics}, pages 2990--2995. IEEE, 2015.

\bibitem[Liu et~al.(2019)Liu, Ott, Goyal, Du, Joshi, Chen, Levy, Lewis, Zettlemoyer, and Stoyanov]{liu2019roberta}
Yinhan Liu, Myle Ott, Naman Goyal, Jingfei Du, Mandar Joshi, Danqi Chen, Omer Levy, Mike Lewis, Luke Zettlemoyer, and Veselin Stoyanov.
\newblock Roberta: A robustly optimized bert pretraining approach.
\newblock \emph{arXiv preprint arXiv:1907.11692}, 2019.

\bibitem[Loshchilov and Hutter(2017)]{loshchilov2017decoupled}
Ilya Loshchilov and Frank Hutter.
\newblock Decoupled weight decay regularization.
\newblock \emph{arXiv preprint arXiv:1711.05101}, 2017.

\bibitem[Lu and Grauman(2013)]{lu2013story}
Zheng Lu and Kristen Grauman.
\newblock Story-driven summarization for egocentric video.
\newblock In \emph{Proceedings of the IEEE conference on computer vision and pattern recognition}, pages 2714--2721, 2013.

\bibitem[Mahasseni et~al.(2017)Mahasseni, Lam, and Todorovic]{mahasseni2017unsupervised}
Behrooz Mahasseni, Michael Lam, and Sinisa Todorovic.
\newblock Unsupervised video summarization with adversarial lstm networks.
\newblock In \emph{Proceedings of the IEEE conference on Computer Vision and Pattern Recognition}, pages 202--211, 2017.

\bibitem[Mishra et~al.(2022)Mishra, Diwan, Srinivasa, and Srinivasaraghavan]{mishra2022semi}
Prakhar Mishra, Chaitali Diwan, Srinath Srinivasa, and G Srinivasaraghavan.
\newblock A semi-automatic approach for generating video trailers for learning pathways.
\newblock In \emph{Artificial Intelligence in Education. Posters and Late Breaking Results, Workshops and Tutorials, Industry and Innovation Tracks, Practitioners’ and Doctoral Consortium: 23rd International Conference, AIED 2022, Durham, UK, July 27--31, 2022, Proceedings, Part II}, pages 302--305. Springer, 2022.

\bibitem[Narasimhan et~al.(2021)Narasimhan, Rohrbach, and Darrell]{narasimhan2021clip}
Medhini Narasimhan, Anna Rohrbach, and Trevor Darrell.
\newblock Clip-it! language-guided video summarization.
\newblock \emph{Advances in Neural Information Processing Systems}, 34:\penalty0 13988--14000, 2021.

\bibitem[Narasimhan et~al.(2022)Narasimhan, Nagrani, Sun, Rubinstein, Darrell, Rohrbach, and Schmid]{narasimhan2022tl}
Medhini Narasimhan, Arsha Nagrani, Chen Sun, Michael Rubinstein, Trevor Darrell, Anna Rohrbach, and Cordelia Schmid.
\newblock Tl; dw? summarizing instructional videos with task relevance and cross-modal saliency.
\newblock In \emph{European Conference on Computer Vision}, pages 540--557. Springer, 2022.

\bibitem[Panda and Roy-Chowdhury(2017)]{panda2017collaborative}
Rameswar Panda and Amit~K Roy-Chowdhury.
\newblock Collaborative summarization of topic-related videos.
\newblock In \emph{Proceedings of the IEEE Conference on computer vision and pattern recognition}, pages 7083--7092, 2017.

\bibitem[Park et~al.(2020)Park, Lee, Kim, and Sohn]{park2020sumgraph}
Jungin Park, Jiyoung Lee, Ig-Jae Kim, and Kwanghoon Sohn.
\newblock Sumgraph: Video summarization via recursive graph modeling.
\newblock In \emph{Computer Vision--ECCV 2020: 16th European Conference, Glasgow, UK, August 23--28, 2020, Proceedings, Part XXV 16}, pages 647--663. Springer, 2020.

\bibitem[Potapov et~al.(2014)Potapov, Douze, Harchaoui, and Schmid]{potapov2014category}
Danila Potapov, Matthijs Douze, Zaid Harchaoui, and Cordelia Schmid.
\newblock Category-specific video summarization.
\newblock In \emph{Computer Vision--ECCV 2014: 13th European Conference, Zurich, Switzerland, September 6-12, 2014, Proceedings, Part VI 13}, pages 540--555. Springer, 2014.

\bibitem[Radford et~al.(2021)Radford, Kim, Hallacy, Ramesh, Goh, Agarwal, Sastry, Askell, Mishkin, Clark, et~al.]{radford2021learning}
Alec Radford, Jong~Wook Kim, Chris Hallacy, Aditya Ramesh, Gabriel Goh, Sandhini Agarwal, Girish Sastry, Amanda Askell, Pamela Mishkin, Jack Clark, et~al.
\newblock Learning transferable visual models from natural language supervision.
\newblock In \emph{International conference on machine learning}, pages 8748--8763. PMLR, 2021.

\bibitem[Rochan and Wang(2019)]{rochan2019video}
Mrigank Rochan and Yang Wang.
\newblock Video summarization by learning from unpaired data.
\newblock In \emph{Proceedings of the IEEE/CVF Conference on Computer Vision and Pattern Recognition}, pages 7902--7911, 2019.

\bibitem[Rochan et~al.(2018)Rochan, Ye, and Wang]{rochan2018video}
Mrigank Rochan, Linwei Ye, and Yang Wang.
\newblock Video summarization using fully convolutional sequence networks.
\newblock In \emph{Proceedings of the European conference on computer vision (ECCV)}, pages 347--363, 2018.

\bibitem[Smeaton et~al.(2006)Smeaton, Lehane, O'Connor, Brady, and Craig]{smeaton2006automatically}
Alan~F Smeaton, Bart Lehane, Noel~E O'Connor, Conor Brady, and Gary Craig.
\newblock Automatically selecting shots for action movie trailers.
\newblock In \emph{Proceedings of the 8th ACM international workshop on Multimedia information retrieval}, pages 231--238, 2006.

\bibitem[Smith et~al.(2017)Smith, Joshi, Huet, Hsu, and Cota]{smith2017harnessing}
John~R Smith, Dhiraj Joshi, Benoit Huet, Winston Hsu, and Jozef Cota.
\newblock Harnessing ai for augmenting creativity: Application to movie trailer creation.
\newblock In \emph{Proceedings of the 25th ACM international conference on Multimedia}, pages 1799--1808, 2017.

\bibitem[Soldan et~al.(2022)Soldan, Pardo, Alc{\'a}zar, Caba, Zhao, Giancola, and Ghanem]{soldan2022mad}
Mattia Soldan, Alejandro Pardo, Juan~Le{\'o}n Alc{\'a}zar, Fabian Caba, Chen Zhao, Silvio Giancola, and Bernard Ghanem.
\newblock Mad: A scalable dataset for language grounding in videos from movie audio descriptions.
\newblock In \emph{Proceedings of the IEEE/CVF Conference on Computer Vision and Pattern Recognition}, pages 5026--5035, 2022.

\bibitem[Song et~al.(2015)Song, Vallmitjana, Stent, and Jaimes]{song2015tvsum}
Yale Song, Jordi Vallmitjana, Amanda Stent, and Alejandro Jaimes.
\newblock Tvsum: Summarizing web videos using titles.
\newblock In \emph{Proceedings of the IEEE conference on computer vision and pattern recognition}, pages 5179--5187, 2015.

\bibitem[Sou{\v{c}}ek and Loko{\v{c}}(2020)]{souvcek2020transnet}
Tom{\'a}{\v{s}} Sou{\v{c}}ek and Jakub Loko{\v{c}}.
\newblock Transnet v2: an effective deep network architecture for fast shot transition detection.
\newblock \emph{arXiv preprint arXiv:2008.04838}, 2020.

\bibitem[Sutskever et~al.(2014)Sutskever, Vinyals, and Le]{sutskever2014sequence}
Ilya Sutskever, Oriol Vinyals, and Quoc~V Le.
\newblock Sequence to sequence learning with neural networks.
\newblock \emph{Advances in neural information processing systems}, 27, 2014.

\bibitem[Vaswani et~al.(2017)Vaswani, Shazeer, Parmar, Uszkoreit, Jones, Gomez, Kaiser, and Polosukhin]{vaswani2017attention}
Ashish Vaswani, Noam Shazeer, Niki Parmar, Jakob Uszkoreit, Llion Jones, Aidan~N Gomez, {\L}ukasz Kaiser, and Illia Polosukhin.
\newblock Attention is all you need.
\newblock \emph{Advances in neural information processing systems}, 30, 2017.

\bibitem[Wang et~al.(2020)Wang, Liu, Puri, and Metaxas]{wang2020learning}
Lezi Wang, Dong Liu, Rohit Puri, and Dimitris~N Metaxas.
\newblock Learning trailer moments in full-length movies with co-contrastive attention.
\newblock In \emph{Computer Vision--ECCV 2020: 16th European Conference, Glasgow, UK, August 23--28, 2020, Proceedings, Part XVIII 16}, pages 300--316. Springer, 2020.

\bibitem[Xu et~al.(2023)Xu, Sun, Li, Shi, Zhu, and Du]{xu2023mh}
Yifang Xu, Yunzhuo Sun, Yang Li, Yilei Shi, Xiaoxiang Zhu, and Sidan Du.
\newblock Mh-detr: Video moment and highlight detection with cross-modal transformer.
\newblock \emph{arXiv preprint arXiv:2305.00355}, 2023.

\bibitem[Zhang et~al.(2016{\natexlab{a}})Zhang, Chao, Sha, and Grauman]{zhang2016summary}
Ke Zhang, Wei-Lun Chao, Fei Sha, and Kristen Grauman.
\newblock Summary transfer: Exemplar-based subset selection for video summarization.
\newblock In \emph{Proceedings of the IEEE conference on computer vision and pattern recognition}, pages 1059--1067, 2016{\natexlab{a}}.

\bibitem[Zhang et~al.(2016{\natexlab{b}})Zhang, Chao, Sha, and Grauman]{zhang2016video}
Ke Zhang, Wei-Lun Chao, Fei Sha, and Kristen Grauman.
\newblock Video summarization with long short-term memory.
\newblock In \emph{Computer Vision--ECCV 2016: 14th European Conference, Amsterdam, The Netherlands, October 11--14, 2016, Proceedings, Part VII 14}, pages 766--782. Springer, 2016{\natexlab{b}}.

\bibitem[Zhang et~al.(2018)Zhang, Grauman, and Sha]{zhang2018retrospective}
Ke Zhang, Kristen Grauman, and Fei Sha.
\newblock Retrospective encoders for video summarization.
\newblock In \emph{Proceedings of the European conference on computer vision (ECCV)}, pages 383--399, 2018.

\bibitem[Zhang et~al.(2023)Zhang, Kang, Hooi, Yan, and Feng]{zhang2023deep}
Yifan Zhang, Bingyi Kang, Bryan Hooi, Shuicheng Yan, and Jiashi Feng.
\newblock Deep long-tailed learning: A survey.
\newblock \emph{IEEE Transactions on Pattern Analysis and Machine Intelligence}, 2023.

\bibitem[Zhou et~al.(2018)Zhou, Qiao, and Xiang]{zhou2018deep}
Kaiyang Zhou, Yu Qiao, and Tao Xiang.
\newblock Deep reinforcement learning for unsupervised video summarization with diversity-representativeness reward.
\newblock In \emph{Proceedings of the AAAI Conference on Artificial Intelligence}, 2018.

\end{thebibliography}
}


\end{document}